\documentclass[runningheads]{llncs}
\usepackage[T1]{fontenc}
\usepackage{graphicx}
\usepackage[table]{xcolor}
\usepackage{subcaption}
\usepackage{amsmath}
\usepackage{amssymb}
\usepackage{tabularx}
\usepackage{booktabs}
\usepackage{array}
\usepackage{multirow} 
\usepackage{algpseudocode}
\usepackage{xcolor}      
\usepackage[hidelinks]{hyperref}  
\hypersetup{
  colorlinks=true,        
  citecolor=green         
}
\definecolor{lightgreen}{rgb}{0.3, 0.7, 0.3}
\algrenewcommand\algorithmiccomment[1]{\textcolor{lightgreen}{// #1}}

\usepackage{graphicx,verbatim}
\usepackage{algorithm}
\usepackage{algpseudocode}
\usepackage[table,xcdraw]{xcolor}
\usepackage[T1]{fontenc}
\usepackage{amsmath} 
\usepackage{amssymb}
\usepackage{booktabs}
\usepackage{tabularx}
\usepackage{multirow}
\usepackage{array}

\newcolumntype{Y}{>{\centering\arraybackslash}X}
\newcolumntype{C}{>{\centering\arraybackslash}X}
\usepackage{graphicx,verbatim}

\begin{document}

\title{Modeling Clinical Workflow for SYNTAX Scoring from Coronary Angiography Videos}
%
%
\author{Suzhong Fu\inst{1,2} \and
Jingqi Dong\inst{1,2} \and Xuan Ding\inst{1,2} \and Rui Sun\inst{1,2} \and Yiming Yang\inst{1,2} \and Shuguang Cui\inst{2,1} \and
Zhen Li\inst{2,1}\thanks{Corresponding author: Zhen Li, email: lizhen@cuhk.edu.cn}}
\authorrunning{S. Fu et al.}
%
\institute{FNii-Shenzhen, The Chinese University of Hong Kong, Shenzhen, China \and
School of Science and Engineering, The Chinese University of Hong Kong, Shenzhen, China }
\maketitle              
\begin{abstract}
The SYNTAX score is a clinically established tool for assessing anatomical lesion complexity in coronary artery disease and guiding subsequent treatment. However, automated SYNTAX scoring is commonly formulated as a direct regression problem from coronary angiography videos to patient-level scores. In this work, we reformulate SYNTAX scoring as a vessel segment identity-preserving anatomical reasoning problem and propose a hierarchical modeling framework that explicitly aligns learning with the clinical workflow. Our approach maintains vessel segment identity across frames and views, estimates stenosis severity at the segment level, and aggregates evidence hierarchically according to coronary anatomy. Simultaneously, to address the scarcity of domain-specific data, we integrate and complete multiple public coronary angiography datasets, constructing a large-scale resource featuring completed vessel segmentation and derived structural annotations. Experiments demonstrate that vessel segment-level stenosis embedding enhances explanatory power and reduces prediction variability compared to baseline models, with the $R^2$ score improving by 0.201 and dev STD decreasing by 18.4\%. These results highlight the necessity of structure-aligned modeling for reliable and stable automated SYNTAX scoring from multi-view coronary angiography videos. The GitHub link is \href{https://github.com/VersaceSu7/SYNTAX_score_777}{SYNTAX\_score\_777}.
\end{abstract}

\section{Introduction}

Coronary artery disease (CAD)\cite{mccullough2007coronary,shahjehan2024coronary} remains a leading cause of morbidity and mortality worldwide, and invasive coronary angiography is the clinical gold standard for visualizing coronary lesions and guiding subsequent treatment. The SYNTAX scoring\cite{serruys2009assessment,sianos2005syntax} outlined in clinical guidelines is inherently structured: cardiologists reason over anatomically defined vessel segments across multiple views and integrate segment-level findings hierarchically into left/right and patient-level scores.

\begin{figure}[t]
\includegraphics[width=\textwidth]{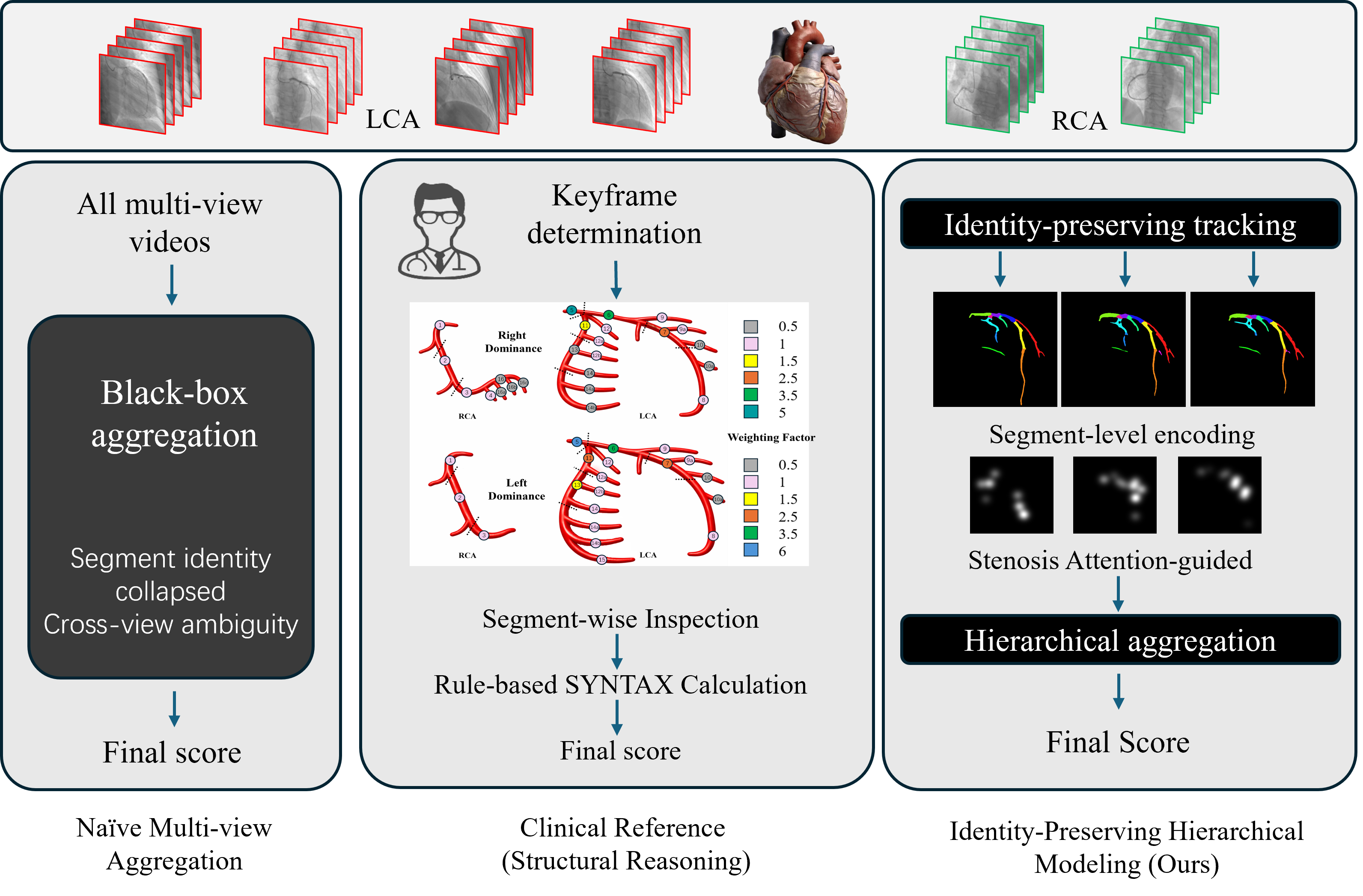}
\caption{Conceptual comparison between naïve multi-view aggregation and identity-preserving hierarchical modeling for automated SYNTAX prediction. Left: Direct aggregation of multi-view features leads to anatomical identity collapse and structural ambiguity. Middle: Clinical reasoning operates on anatomically defined vessel segments and hierarchical integration. Right: The proposed framework preserves segment identity across frames and views, enabling structure-aware aggregation for patient-level SYNTAX prediction.} \label{fig1}
\end{figure}

Despite recent advances in deep learning for coronary angiography analysis, most existing approaches address isolated components of this reasoning process, such as vessel segmentation\cite{fu2025vessam,popov2024dataset,cervantes2019automatic,ma2021self,gao2025spatio,he2025conditional}, stenosis detection\cite{avram2023cathai,labrecque2024evaluation,li2024stqd,pang2021stenosis,mahmoudi2025x}, or dominance classification\cite{kruzhilov2025coronarydominance,akyurek2025real}. A limited number of studies attempt direct end-to-end SYNTAX score prediction from angiographic videos\cite{ponomarchuk2025cardiosyntax}. However, these methods typically formulate the task as a black-box regression from high-dimensional spatiotemporal inputs to a low-dimensional clinical score, collapsing structured anatomical reasoning into flat feature representations. As a result, vessel identity correspondence across views, segment-level lesion attribution, and hierarchical aggregation are not explicitly modeled, limiting interpretability, robustness, and generalization across heterogeneous clinical data.

In this work, we argue that automated SYNTAX prediction should be formulated as a structured anatomical reasoning problem rather than a direct regression task. We propose a clinically inspired, vessel segment identity-preserving framework that explicitly models anatomical identity propagation across frames, segment-level lesion estimation, and hierarchical aggregation across vessels and views. Central to our approach is a segment-aware decision field that embeds discrete segment-level stenosis assessments into a continuous, topology-aware representation, enabling spatially coherent reasoning while preserving anatomical correspondence throughout the modeling pipeline.

Simultaneously, to address the scarcity of domain-specific data, we integrate and complete multiple public coronary angiography datasets, constructing a large-scale resource featuring completed vessel segmentation and derived structural annotations. Experiments demonstrate that explicitly modeling anatomical structure and hierarchical aggregation leads to consistently improved patient-level SYNTAX score prediction performance and reduced prediction variability compared with strong baselines, with particularly pronounced benefits in anatomically complex coronary systems. Our contributions are summarized as follows:
\begin{enumerate}
    \item Introduce a vessel segment identity-preserving modeling framework that aligns learning objectives with the degree of stenosis in each coronary vessel segment;
    \item Construct a coronary artery representation network capable of jointly learning from vessel segment stenosis Attention Heatmaps and multi-view videos to predict SYNTAX scores;
    \item Integrate and refine commercially available high-quality coronary X-ray data, utilizing progressive generation to complete segmentation labels for all 1.2M images;
    \item  Experiments validate that our framework yields more stable and interpretable SYNTAX predictions across heterogeneous multi-view data.
\end{enumerate}

\section{Method}
\label{sec:method}


\textbf{Problem Formulation under Hierarchical Structure.}
A coronary angiography examination consists of multiple videos acquired from heterogeneous viewing angles, each containing hundreds of frames. Formally, an examination is represented as $\mathcal{V} = \{V_1, \dots, V_N\}$, where each video $V_i = \{I_{i,1}, \dots, I_{i,T_i}\}$ captures spatiotemporal vessel dynamics.
Unlike conventional video regression problems, SYNTAX scoring is inherently hierarchical. The final clinical targets ($y_{\text{left}}$, $y_{\text{right}}$, $y_{\text{full}}$) are determined by localized lesion decisions over anatomically defined vessel segments, which are further organized within left and right coronary systems. Therefore, the learning objective is not merely to approximate a mapping $F: \mathbb{R}^{N \times T_i \times H \times W} \rightarrow \mathbb{R}^3,$ but to construct intermediate representations that preserve segment-level identity and hierarchical structure before aggregation.
This perspective motivates explicit enforcement of anatomical identity and structured feature integration as learning constraints, rather than direct end-to-end regression.
Because raw angiography videos include non-informative phases caused by absent or dissipated contrast agent and cardiac motion, the model is trained on high-quality clips rather than uniformly sampling all frames. We annotated high-quality video segments from all frame-level data of 120 patients and trained a binary ResNet-34 classifier using more than 800 videos to distinguish informative frames from other frames. The complete supplementary videos are kept untrimmed for transparency, whereas training and heatmap construction use the selected high-quality clips.

\begin{figure}[t]
\includegraphics[width=\textwidth]{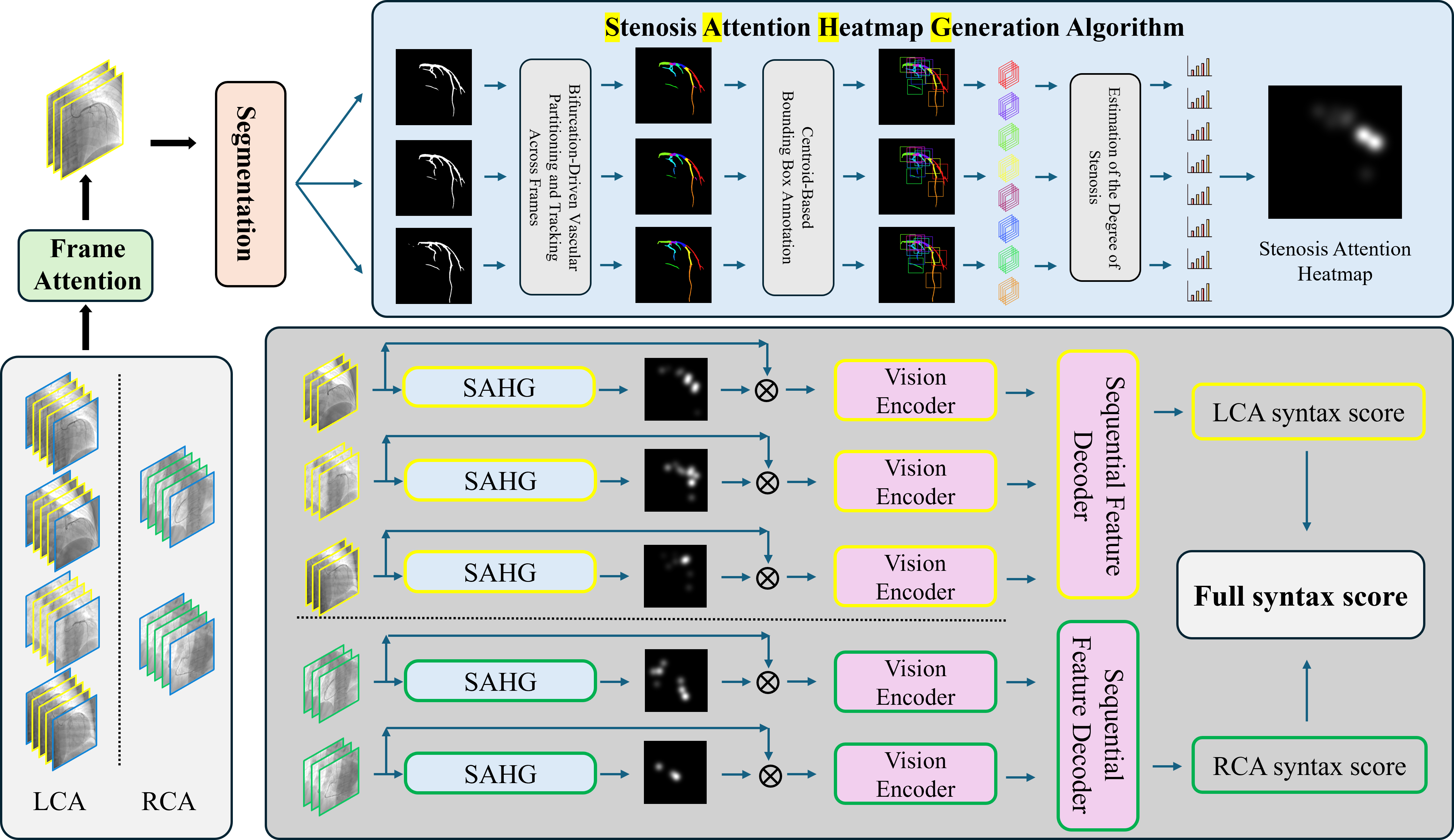}
\caption{Schematic diagram of the proposed syntax score prediction framework for a single case. Key frames are selected from the patient's videos to ensure the accuracy of feature information sources. High-quality frames from each perspective video serve as sources for generating narrow attention heatmaps. The stenosis attention heatmap is combined with video key frames as joint inputs to the network, comprising a video-based visual encoder and a sequential feature decoder that aggregates multi-video features for syntax prediction.} \label{fig2}
\end{figure}

\textbf{Identity-Preserving Anatomical Modeling.}
To prevent segment anatomical identity collapse across frames and views, we represent vessel segments as topology-defined structural entities rather than appearance-dependent pixel regions.
In coronary angiography, pixel-level representations are highly sensitive to projection angle, vessel foreshortening, and contrast dynamics, causing the same anatomical segment to exhibit drastically different appearances across views.
Given a vessel segmentation mask $M$, we extract its centerline and represent the vascular tree as a connectivity graph $G = (\mathcal{N}, \mathcal{E})$, where nodes correspond to endpoints or branching points and edges correspond to centerline branches.
By decomposing the graph at bifurcation nodes, we obtain a set of linear vessel segments $\{S_k\}_{k=1}^{K}$, each corresponding to an anatomically defined coronary segment.
This topology-based decomposition transforms the vascular tree into structurally stable segment entities whose identities remain consistent across frames and views, despite variations in motion, contrast injection, and imaging projection.
By anchoring segment identity to vascular topology rather than pixel appearance, the model retains anatomically meaningful correspondence throughout subsequent temporal propagation and multi-view aggregation.
The segment extraction follows the skeleton-graph splitting strategy in VesSAM~\cite{fu2025vessam}: bifurcation and endpoint nodes are first detected on the vessel skeleton, and connected skeleton branches are then used to define segment candidates. These segments are mainly used to produce stable, non-overlapping video crops for downstream stenosis estimation. To reduce error accumulation in this multi-stage pipeline, segmentation masks are iteratively refined, segment tracking is performed at the frame level, and highly overlapping vessel crops are removed before stenosis estimation.

\textbf{Segment-Aware Decision Field.}
To preserve both the magnitude and anatomical support of segment-level clinical findings, we introduce a \emph{segment-aware decision field}, which embeds discrete stenosis assessments into a continuous spatial representation while maintaining explicit segment correspondence.
For each tracked vessel segment $S_k$, we estimate a stenosis severity score $s_k$ from its corresponding vessel crop using the segment-level stenosis estimator, and map it to a non-negative modulation coefficient $\alpha_k$.
The spatial contribution of segment $k$ at location $p \in \mathbb{R}^2$ is modeled as
\[
H_k(p) = \alpha_k \exp\left(
-\frac{1}{2}(p - \mu_k)^\top \Sigma_k^{-1} (p - \mu_k)
\right),
\]
where $\mu_k$ denotes the spatial center of segment $S_k$, and $\Sigma_k$ encodes its spatial extent and orientation.
In practice, $(\mu_k, \Sigma_k)$ are derived from the geometry of the tracked segment (e.g., centerline statistics) and are fixed given the segmentation and tracking results.
The final decision field is obtained by superposing contributions from all segments,
$H(p) = \sum_{k=1}^{K} H_k(p),$
resulting in a continuous spatial field that encodes the distribution of segment-level stenosis evidence over the vascular tree. We do not claim the Gaussian form is optimal; it serves as a continuous carrier of discrete clinical decisions.

This construction embeds discrete segment-level clinical decisions into a continuous, topology-aware representation that preserves both anatomical locality and segment identity.
Unlike conventional attention or saliency maps, which are typically learned from appearance features and vary across views, the proposed decision field is explicitly anchored to tracked anatomical segments and remains consistent across frames and views.
As a result, the decision field enforces identity-preserving modulation during downstream feature learning and hierarchical aggregation, rather than serving as a view-specific importance weighting.
This superposition also makes the representation less sensitive to occasional pseudo-bifurcation points caused by 2D vessel overlap: even if a projected vessel branch is split into multiple crops, the local stenosis evidence is still mapped back to the same image support and contributes to the final decision field.

\textbf{Hierarchical Multi-Video Aggregation.}
Case-level SYNTAX prediction requires integrating evidence across multiple angiographic videos while preserving vessel-level correspondence.
In multi-view examinations, different videos often capture distinct subsets of coronary segments under heterogeneous projections.
Directly pooling video-level embeddings implicitly assumes that such embeddings are anatomically aligned, an assumption that is violated when segment identity is not explicitly enforced.

We therefore adopt a hierarchical aggregation strategy aligned with coronary anatomy, rather than performing direct, view-agnostic pooling.
Each video is first encoded into a video-level representation using a feature extractor that is explicitly conditioned on the segment-aware decision field.
The decision field serves as an identity-preserving modulation, emphasizing anatomically localized stenosis evidence while suppressing appearance-driven ambiguity.

Video-level representations are then grouped according to their associated coronary system (left or right), reflecting the anatomical organization underlying SYNTAX scoring, before subsequent aggregation.
This hierarchy-aware aggregation prevents feature mixing between anatomically distinct coronary systems, ensuring that evidence from the left and right coronary arteries is not inadvertently entangled.
By enforcing vessel-level correspondence prior to patient-level integration, the model preserves the anatomical structure required for reliable SYNTAX score estimation and mitigates instability caused by heterogeneous multi-view inputs.

\section{Experiments}

\textbf{Datasets.} Data barriers remain a significant obstacle to progress in this field, as most current research relies on proprietary datasets. No single dataset simultaneously contains large-scale multi-view videos, complete vessel segmentation, stenosis attributes, and patient-level SYNTAX scores. This fragmented supervision setting introduces structural inconsistency and increases the risk of anatomical identity collapse during learning. To enable structure-aware learning under this fragmented supervision regime, we utilized high-quality annotations from DCA1 and XCAD to complete missing segmentation labels in the CardiaSyntax and SYNTAX-Score datasets through progressive segmentation refinement and structural annotation transfer. Additionally, segmentation labels for the ARCADE dataset were optimized. The high-quality clip annotation set used for frame selection contains 120 patients and more than 800 videos, and will be released together with the code and trained weights. Dataset characteristics and segmentation availability are summarized in Table~\ref{tab:datasets}.

\begin{table}[t]
\centering
\caption{Summary of datasets used in this study.
Segmentation availability is reported for original annotations, optimized and after completion by our pipeline.}
\label{tab:datasets}
\begin{tabularx}{\linewidth}{YYYYYY}
\hline
\textbf{Attribute} 
& \textbf{Syntax-score\cite{mahmoudi2025x}} 
& \textbf{Cardio-Syntax\cite{ponomarchuk2025cardiosyntax}} 
& \textbf{ARCADE\cite{popov2024dataset}} 
& \textbf{DCA1\cite{cervantes2019automatic}} 
& \textbf{XCAD\cite{ma2021self}} \\ \hline
Dim & 2D & 3D & 2D & 2D & 2D \\
Cases|Imgs & 237|3459 & 3018|1.2M & *|1500 & *|130 & *|126 \\
Multi-view & \checkmark & \checkmark & $\times$ & $\times$ & $\times$ \\
Dominance & \checkmark & 1025 & $\times$ & $\times$ & $\times$ \\
SYNTAX & \checkmark & 1844 & $\times$ & $\times$ & $\times$ \\
Segmentation & \textbf{\textcolor{blue!70}{Completed}} & \textbf{\textcolor{blue!70}{Completed}} &\textbf{\textcolor{blue!70}{Optimized}}  & Original & Original \\
Stenosis & \checkmark & $\times$ & \checkmark & $\times$ & $\times$ \\\hline
\end{tabularx}
\end{table}

\textbf{Baselines.}
We reproduce the prior state-of-the-art method (CarS)\cite{ponomarchuk2025cardiosyntax} under identical 5-fold patient-level splits.
To isolate structural contributions, we evaluate:
(i) CarS + SAH, which augments the baseline with segment-aware hierarchical guidance,
(ii) Ours w/o SAH, which removes hierarchical guidance from our framework, and
(iii) Ours, the complete identity-preserving model.
We focus on CarS because it is the directly reproducible temporal coronary angiography benchmark for SYNTAX regression. CathAI-like systems target 2D angiographic stenosis detection and severity estimation rather than full-pipeline video-based SYNTAX scoring, and their code, weights, and datasets are not publicly available for direct adaptation.

\textbf{Evaluation}
Performance is assessed from three complementary perspectives:
(1) explanatory power measured by $R^2$,
(2) prediction bias measured by mean and median error, and
(3) prediction dispersion measured by the standard deviation of prediction errors (STD).
Unless otherwise specified, values reported in the main tables correspond to mean $\pm$ standard deviation across folds.
These regression metrics are used to evaluate prediction reliability and to allow direct comparison with CardioSyntax~\cite{ponomarchuk2025cardiosyntax}. We acknowledge that clinically risk-stratified metrics based on SYNTAX score thresholds are also important for downstream decision support and should be included in future external clinical validation.

\section{Results and analysis}

Table~\ref{tab:overall_syntax} summarizes patient-level SYNTAX score prediction performance on the test dataset.
Compared with the baseline, structure-aware models consistently improve explanatory power while simultaneously reducing prediction variability.
These gains directly reflect the benefit of explicitly preserving anatomical identity and hierarchical structure during multi-view aggregation.

The complete model achieves the highest $R^2$ for both left (0.469) and full (0.529) SYNTAX scores.
Notably, performance gains are more pronounced for the left coronary system, which is characterized by more complex branching topology and a larger number of anatomically distinct segments.
This trend is consistent with our formulation in Section~2, indicating that identity-preserving and topology-aware modeling is particularly advantageous under higher anatomical complexity.
Structure-aware variants exhibit increased mean bias for the full SYNTAX score, indicating sensitivity to global score scaling.

Given that SYNTAX scoring is itself derived from discrete, segment-level clinical decisions, improved variance explanation with near-zero median bias reflects enhanced structural consistency rather than random prediction drift.
As shown in Tables~\ref{tab:overall_syntax} and~\ref{tab:left_right_full}, STD consistently decreases for structure-aware models, with particularly pronounced reductions for the left and full SYNTAX scores.
This reduction supports the claim that explicitly preserving anatomical identity mitigates variability introduced by heterogeneous multi-view aggregation.

\begin{table}[t]
\centering
\caption{Overall SYNTAX score prediction performance on the full test dataset.
Mean $\pm$ standard deviation over 5-fold patient-level cross-validation are reported.}
\label{tab:overall_syntax}
\begin{tabularx}{\linewidth}{l*{4}{Y}}
\hline
Model & $R^2 \uparrow$ & Bias Mean & Bias Median & STD $\downarrow$ \\
\hline
CarS 
& 0.328 $\pm$ 0.035 
& $-0.41 \pm 1.56$ 
& $0.01 \pm 0.02$ 
& 8.36 $\pm$ 0.22 \\

CarS w SAH 
& 0.420 $\pm$ 0.023 
& $-1.05 \pm 0.32$ 
& $0.04 \pm 0.08$ 
& 7.83 $\pm$ 0.16 \\

Ours w/o SAH 
& 0.467 $\pm$ 0.047 
& $-2.11 \pm 1.19$ 
& $0.06 \pm 0.12$ 
& 7.17 $\pm$ 0.49 \\

\textbf{Ours} 
& \textbf{0.529 $\pm$ 0.027} 
& $-1.96 \pm 0.62$ 
& $0.02 \pm 0.03$ 
& \textbf{6.82 $\pm$ 0.09} \\

$\Delta$ vs. CarS 
& \textbf{\textcolor{blue!70}{+0.201}} 
& \textcolor{blue!70}{-1.55} 
& \textcolor{blue!70}{+0.01} 
& \textbf{\textcolor{blue!70}{-1.54(-18.4\%)}} \\
\hline
\end{tabularx}
\end{table}

\begin{table}[t]
\centering
\caption{Left and right coronary SYNTAX prediction performance on the full dataset.
Mean values over 5-fold patient-level cross-validation are reported.
}
\label{tab:left_right_full}
\begin{tabularx}{\linewidth}{lYYY YYY}
\hline
\multicolumn{4}{c}{\textbf{Left coronary (LCA)}} & \multicolumn{3}{c}{\textbf{Right coronary (RCA)}} \\
\hline
Model & $R^2 \uparrow$ & Bias Mean  & STD $\downarrow$
& $R^2 \uparrow$ & Bias Mean  & STD $\downarrow$ \\
\hline
CarS
& 0.164 & 0.36 & 7.71
& \textbf{0.253} & -0.77 & 2.45 \\

CarS w SAH
& 0.310 & -0.38 & 7.14
& 0.239 & -0.67 & 2.47 \\

Ours w/o SAH
& 0.422 & -1.40 & 6.29
& 0.232 & -0.71 & 2.50 \\

\textbf{Ours}
& \textbf{0.470} & -1.15 & \textbf{6.14}
& 0.254 & -0.81 & \textbf{2.44} \\

$\Delta$ vs. CarS
& \textbf{\textcolor{blue!70}{+0.306}} 
& \textcolor{blue!70}{-1.51} 
& \textbf{\textcolor{blue!70}{-1.57}}
& \textcolor{blue!70}{+0.001} 
& \textcolor{blue!70}{-0.04} 
& \textbf{\textcolor{blue!70}{-0.01}} \\
\hline
\end{tabularx}
\end{table}

\begin{table}[t]
\centering
\caption{SYNTAX prediction performance on non-zero cases.
Mean values over 5-fold patient-level cross-validation are reported.}
\label{tab:nonzero_syntax}
\begin{tabularx}{\linewidth}{lCCCC}
\hline
 & \multicolumn{2}{c}{\textbf{Left SYNTAX}} & \multicolumn{2}{c}{\textbf{Overall SYNTAX}} \\
\textbf{Method}
& $R^2 \uparrow$ & STD $\downarrow$
& $R^2 \uparrow$ & STD $\downarrow$ \\
\hline
CarS
& -0.338$\pm$0.160 & 10.914$\pm$0.512
& -0.186$\pm$0.063 & 11.786$\pm$0.418 \\

CarS w SAH
& -0.096$\pm$0.031 & 10.249$\pm$0.169
& -0.015$\pm$0.042 & 10.983$\pm$0.275 \\

Ours w/o SAH
& 0.101$\pm$0.107 & \textbf{8.456$\pm$0.959}
& 0.089$\pm$0.055 & 9.155$\pm$0.917 \\

\textbf{Ours}
& \textbf{0.160$\pm$0.033} & 8.500$\pm$0.429
& \textbf{0.179$\pm$0.052} & \textbf{8.978$\pm$0.312} \\

$\Delta$ vs. CarS
& \textbf{\textcolor{blue!70}{+0.498}}
& \textbf{\textcolor{blue!70}{-2.41}}
& \textbf{\textcolor{blue!70}{+0.365}}
& \textbf{\textcolor{blue!70}{-2.81}} \\
\hline
\end{tabularx}
\end{table}

\textbf{Challenge SYNTAX cases experiments.} We further evaluate model performance on non-zero SYNTAX cases, which represent a more challenging clinical setting.
As shown in Table~\ref{tab:nonzero_syntax}, overall performance decreases across all methods, reflecting the increased difficulty of this subset.
Nevertheless, structure-aware models consistently improve both explanatory power and prediction stability compared with the baseline.
In particular, the complete identity-preserving model achieves higher $R^2$ while simultaneously reducing deviation standard deviation for both left and overall SYNTAX scores, indicating more stable patient-level predictions under harder conditions.
This suggests that explicitly modeling anatomical identity and hierarchical aggregation mitigates performance degradation when trivial cases are removed.

\begin{table}[t]
\centering
\caption{Effect of removing multi-view aggregation on SYNTAX prediction.
Mean $R^2$ values across 5-fold patient-level cross-validation are reported.}
\label{tab:ablation_no_multiview}
\begin{tabularx}{\linewidth}{lCCC}
\hline
\textbf{Setting} & \textbf{Left} & \textbf{Right} & \textbf{Full} \\
\hline
Mean pooling (single-view) & -4.34$\pm$5.97 & 0.07$\pm$0.15 & -4.47$\pm$2.99 \\
Mean pooling + SAH & -5.13$\pm$5.61 & -0.12$\pm$0.24 & -5.34$\pm$7.68 \\
\hline
\end{tabularx}
\end{table}

\textbf{Effect of Removing Multi-view Aggregation.}
We further analyze a degenerate setting in which multi-view aggregation is removed and SYNTAX prediction is performed using single-view representations.
As shown in Table~\ref{tab:ablation_no_multiview}, this setting leads to severely degraded performance, with consistently negative $R^2$ values for left and full SYNTAX scores.
Notably, incorporating heatmap-based guidance without multi-view aggregation does not improve performance and often further destabilizes prediction.
This observation indicates that reliable SYNTAX prediction fundamentally requires explicit aggregation of segment-level evidence across multiple angiographic views.

\section{Conclusion}
In this work, we proposed a vessel segment identity-preserving hierarchical framework for automated SYNTAX score prediction from coronary angiography videos, explicitly aligning learning with the structured clinical workflow underlying SYNTAX scoring. By maintaining vessel segment identity across frames and views and aggregating evidence according to the stenosis severity of coronary anatomy, our method avoids anatomical identity collapse inherent in direct end-to-end regression. We integrated and completed multiple public coronary angiography datasets to alleviate the current data constraints in this field. Experiments across multiple datasets demonstrate consistently improved explanatory power and reduced prediction variability compared with strong baselines, with particularly pronounced gains for left coronary systems and non-zero SYNTAX cases. These results highlight the importance of structure-aware and topology-informed modeling for reliable angiographic video analysis. Our current study is still limited by the absence of an independent external clinical cohort, the lack of risk-class-wise clinical evaluation, the relatively narrow comparison set, and the use of a Gaussian decision field as a practical continuous representation rather than an optimized spatial kernel. Future work will further validate the framework on external cohorts and explore more clinically grounded spatial representations. Our work provides a step toward more clinically faithful and robust automated assessment of coronary disease complexity.

\begin{credits}
\subsubsection{\ackname} This work was supported in part by Guangdong S\&T Programme with Grant No. 2024B0101030002, the Basic Research Project No. HZQB-KCZYZ-2021067 of Hetao Shenzhen-HK ST Cooperation Zone, the Shenzhen Outstanding Talents Training Fund 202002, the NSFC with Grant No. 62293482, by NSFC with Grant No. 62573371, by the Guangdong Province Radio Science Data Center with grant No. 2025B1212070001, by the Shenzhen General Program No. JCYJ202205301436 00001, by the Guangdong Research Project No. 2017ZT07X152 and No. 2019CX01X104, by the Guangdong Provincial Key Laboratory of Future Networks of Intelligence (Grant No. 2022B1212010001), by the NSFC 61931024 \& 12326610, by the Shenzhen Key Laboratory of Big Data and Artificial Intelligence (Grant No. SYSPG20241211173853027), by National Key Research and Development Program of China2025YFF0515300 and 2025YFF0515304, the Open Project Program (Grant No. QHSFCS- 2606) of Key Laboratory of Tibetan Information Processing, Ministry of Education, by the Shenzhen-Hong Kong Joint Funding No. SGDX20211123112401002, and by Tencent \& Huawei Open Fund 2024E0009 \& 202301030019.

\subsubsection{Declaration of Competing Interest}

The authors have no competing interests to declare that are relevant to the content of this article.

\end{credits}
%
%
%

\begin{thebibliography}{8}
\bibitem{mccullough2007coronary}
McCullough, Peter A.: Coronary artery disease. Clinical Journal of the American Society of Nephrology 2.3 (2007): 611-616.

\bibitem{shahjehan2024coronary}
Shahjehan, Rai Dilawar, Sanjeev Sharma, and Beenish S. Bhutta.: Coronary artery disease. StatPearls [Internet]. StatPearls Publishing, 2024.

\bibitem{serruys2009assessment}
Serruys, Patrick W., et al.: Assessment of the SYNTAX score in the Syntax study. EuroIntervention 5.1 (2009): 50-56.

\bibitem{sianos2005syntax}
Sianos, Georgios, et al.: The SYNTAX Score: an angiographic tool grading the complexity of coronary artery disease. EuroIntervention 1.2 (2005): 219-227.

\bibitem{fu2025vessam}
Fu, Suzhong, et al.: VesSAM: Efficient Multi-Prompting for Segmenting Complex Vessel. 2025 IEEE International Conference on Bioinformatics and Biomedicine (BIBM). IEEE, 2025.

\bibitem{popov2024dataset}
Popov, Maxim, et al.: Dataset for automatic region-based coronary artery disease diagnostics using X-ray angiography images. Scientific data 11.1 (2024): 20.

\bibitem{cervantes2019automatic}
Cervantes-Sanchez, Fernando, et al.: Automatic segmentation of coronary arteries in X-ray angiograms using multiscale analysis and artificial neural networks. Applied Sciences 9.24 (2019): 5507.

\bibitem{ma2021self}
Ma, Yuxin, et al.: Self-supervised vessel segmentation via adversarial learning. proceedings of the IEEE/CVF international conference on computer vision. 2021.

\bibitem{he2025conditional}
He, Yanglong, et al.: Conditional Virtual Imaging for Few-Shot Vascular Image Segmentation. IEEE Transactions on Medical Imaging (2025).

\bibitem{gao2025spatio}
Gao, Yunlong, et al.: Spatio-Temporal correspondence attention network for vessel segmentation in X-ray coronary angiography. Biomedical Signal Processing and Control 99 (2025): 106792.

\bibitem{avram2023cathai}
Avram, Robert, et al.: CathAI: fully automated coronary angiography interpretation and stenosis estimation. npj Digital Medicine 6.1 (2023): 142.

\bibitem{labrecque2024evaluation}
Labrecque Langlais, Élodie, et al.: Evaluation of stenoses using AI video models applied to coronary angiography. NPJ digital medicine 7.1 (2024): 138.

\bibitem{li2024stqd}
Li, Xinyu, et al.: STQD-Det: Spatio-temporal quantum diffusion model for real-time coronary stenosis detection in X-ray angiography. IEEE Transactions on Pattern Analysis and Machine Intelligence 46.12 (2024): 9908-9920.

\bibitem{pang2021stenosis}
Pang, Kun, et al.: Stenosis-DetNet: Sequence consistency-based stenosis detection for X-ray coronary angiography. Computerized Medical Imaging and Graphics 89 (2021): 101900.

\bibitem{kruzhilov2025coronarydominance}
Kruzhilov, Ivan, et al.: CoronaryDominance: Angiogram dataset for coronary dominance classification. Scientific Data 12.1 (2025): 341.

\bibitem{akyurek2025real}
Akyürek, Hasan Ali.: Real-Time Coronary Artery Dominance Classification from Angiographic Images Using Advanced Deep Video Architectures. Diagnostics 15.10 (2025): 1186.

\bibitem{ponomarchuk2025cardiosyntax}
Ponomarchuk, Alexander, et al.: CardioSyntax: End-to-End SYNTAX Score Prediction-Dataset, Benchmark and Method. 2025 IEEE/CVF Winter Conference on Applications of Computer Vision (WACV). IEEE, 2025.

\bibitem{mahmoudi2025x}
Mahmoudi, Seyed Sajjad, et al.: X-ray coronary angiogram images and SYNTAX score to develop machine-learning algorithms for CHD diagnosis. Scientific Data 12.1 (2025): 471.



\end{thebibliography}
%

\end{document}